\documentclass[runningheads]{llncs}
\usepackage{graphicx}
\usepackage{booktabs}
\usepackage{tabularx}
\usepackage{tikz}
\usepackage{comment}
\usepackage{amsmath,amssymb} 
\usepackage{float}
\usepackage{color}
\usepackage{multirow}

\newcommand{\ie}{{\it i.e.}\hspace{0.1cm}}
\def\etal{\textit{et al.}}

\newcommand*\samethanks[1][\value{footnote}]{\footnotemark[#1]}
\usepackage[hyphens]{url}

\usepackage[accsupp]{axessibility}  

\begin{document}
\pagestyle{headings}
\mainmatter
\def\ECCVSubNumber{3351}  

\title{S$^2$Contact: Graph-based Network for 3D Hand-Object Contact Estimation with Semi-Supervised Learning} 


\titlerunning{S$^2$Contact}
%
\author{Tze Ho Elden Tse\inst{1}\thanks{Equal contribution} \and
Zhongqun Zhang\inst{1}\samethanks \and
Kwang In Kim\inst{2} \and
Ale\u{s} Leonardis\inst{1} \and
Feng Zheng\inst{3} \and
Hyung Jin Chang\inst{1}}
\authorrunning{THE. Tse et al.}
%
\institute{University of Birmingham, UK \and
UNIST, Korea \and
SUSTech, China 
}
\maketitle

\begin{abstract}
Despite the recent efforts in accurate 3D annotations in hand and object datasets, there still exist gaps in 3D hand and object reconstructions. 
Existing works leverage contact maps to refine inaccurate hand-object pose estimations and generate grasps given object models. However, they require explicit 3D supervision which is seldom available and therefore, are limited to constrained settings, e.g., where thermal cameras observe residual heat left on manipulated objects.
In this paper, we propose a novel semi-supervised framework that allows us to learn contact from monocular images. Specifically, we leverage visual and geometric consistency constraints in large-scale datasets for generating pseudo-labels in semi-supervised learning and propose an efficient graph-based network to infer contact. 
Our semi-supervised learning framework achieves a favourable improvement over the existing supervised learning methods trained on data with `limited' annotations.
Notably, our proposed model is able to achieve superior results with less than half the network parameters and memory access cost when compared with the commonly-used PointNet-based approach. We show benefits from using a contact map that rules hand-object interactions to produce more accurate reconstructions. We further demonstrate that training with pseudo-labels can extend contact map estimations to out-of-domain objects and generalise better across multiple datasets. Project page is available.\footnote[1]{\url{https://eldentse.github.io/s2contact/}}
\end{abstract}
\section{Introduction}

\begin{figure}
\centering
\includegraphics[width=0.85\linewidth]{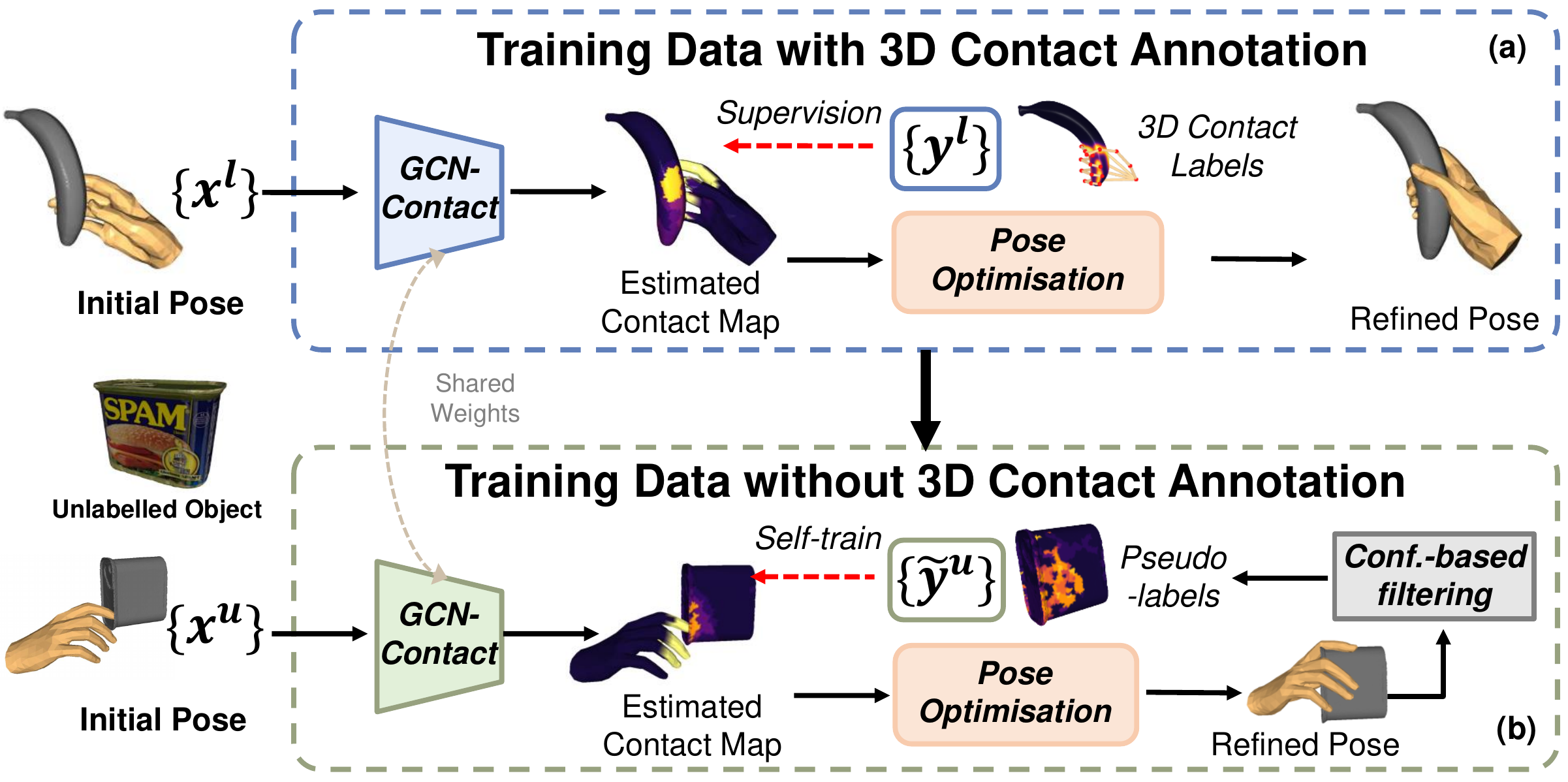}
\caption{
Overview of our semi-supervised learning framework, S$^2$Contact. (a) The model is pre-trained on 
a small annotated dataset. (b) Then, it is deployed on unlabelled datasets to collect pseudo-labels. The pseudo-labels are filtered with confidence-based on visual and geometric consistencies. Upon predicting the contact map, the hand and object poses are jointly optimised to achieve target contact via a contact model~\cite{grady2021contactopt}.
}
\label{fig:highlevel_framework}
\end{figure}

Understanding hand-object interactions have been an active area of study in recent years \cite{hasson2019learning,hasson2020leveraging,hasson20_handobjectconsist,cao2020reconstructing,tekin2019h+,Kwon_2021_ICCV,jiang2021hand,yang2021cpf,liu2021semi,tse2022collaborative}. Besides common practical applications in augmented and virtual reality \cite{han2020megatrack,wang2020rgb2hands,mueller2019real,tse2022collaborative,tse2019no,zheng2022tp}, it is a key ingredient to advanced human-computer interaction \cite{ueda2003hand} and imitation learning in robotics \cite{zhang2018deep}.
In this paper, as illustrated in Figure \ref{fig:highlevel_framework}, we tackle the problem of 3D reconstruction of the hand and manipulated object with the focus on contact map estimation. 


Previous works in hand-object interactions typically formulate this as a joint hand and object pose estimation problem. Along with the development of data collection and annotation methods, more accurate 3D annotations for real datasets \cite{FirstPersonAction_CVPR2018,hampali2020honnotate,chao2021dexycb} are available for learning-based methods \cite{doosti2020hope,tekin2019h+}. Despite the efforts, there still exist gaps between hand-object pose estimation and contact as ground-truth in datasets are not perfect. Recent works attempt to address this problem with interaction constraints (attraction and repulsion) under an optimisation framework \cite{hasson20_handobjectconsist,cao2020reconstructing,yang2021cpf}. However, inferred poses continue to exhibit sufficient error to cause unrealistic hand-object contact, making downstream tasks challenging \cite{grady2021contactopt}. In addition, annotations under constrained laboratory environments rely on strong priors such as limited hand motion which prevents the trained model from generalising to novel scenes and out-of-domain objects.

To address the problem of hand-object contact modelling,
Brahmbhatt~\etal~\cite{brahmbhatt2019contactdb} used thermal cameras observing the heat transfer from hand to object after the grasp to capture detailed ground-truth contact. Their follow-up work contributed a large grasp dataset (\emph{ContactPose}) with contact maps and hand-object pose annotations.
Recent works are able to leverage contact maps to refine inaccurate hand-object pose estimations \cite{grady2021contactopt} and generate grasps given object model \cite{jiang2021hand}. Therefore, the ability to generate an accurate contact map is one of the key elements to reasoning physical contact. However, the number of annotated objects is incomparable to manipulated objects in real life and insufficient to cover a wide range of human intents. Furthermore, obtaining annotations for contact maps is non-trivial as it requires thermal sensors during data collection.

To enable the wider adoption of contact maps, we propose a unified framework that leverages existing hand-object datasets for generating pseudo-labels in semi-supervised learning. Specifically, we propose to exploit the visual and geometric consistencies of contact maps in hand-object interactions. This is built upon the idea that the poses of the hands and objects are highly-correlated where the 3D pose of the hand often indicates the orientation of the manipulated object. We further extend this by enforcing our contact consistency loss for the contact maps across a video.

As the input to contact map estimator are in the form of point clouds, recent related works \cite{grady2021contactopt,jiang2021hand} typically follow a PointNet-based architectures \cite{qi2017pointnet,qi2017pointnet++}. This achieves permutation invariance of points by operating on each point independently and subsequently applying a symmetric function to accumulate features \cite{wang2019dynamic}. However, the network performances are limited as points are treated independently at a local scale to maintain permutation invariance. To overcome this fundamental limitation, many recent approaches adopt graph convolutional networks (GCN) \cite{defferrard2016convolutional,kipf2016semi} and achieve state-of-the-art performances in 3D representation learning on point clouds for classification, part segmentation and semantic segmentation \cite{wang2019dynamic,li2019deepgcns,lin2020convolution}. The ability to capture local geometric structures while maintaining permutation invariance is particularly important for estimating contact maps. However, it comes at the cost of high computation and memory usage for constructing a local neighbourhood with $K$-nearest neighbour ($K$-NN) search on point clouds at each training epoch.
For this reason, we design a graph-based neural network that demonstrates superior results with less than half the learning parameters and faster convergence.


Our contributions are three-fold:
\begin{itemize}
    \setlength{\itemsep}{0pt}%
    \setlength{\parskip}{0pt}%
    \item We propose a novel semi-supervised learning framework that combines pseudo-label with consistency training. Experimental results demonstrate the effectiveness of this training strategy.
    \item We propose a novel graph-based network for processing hand-object point clouds, which is at least two times more efficient than PointNet-based architecture for estimating contact between hand and object.
    \item We conduct comprehensive experiments on three commonly-used hand-object datasets. Experiments show that our proposed framework S$^2$Contact outperforms recent semi-supervised methods.
    
\end{itemize}
\section{Related work}

Our work tackles the problem of hand and object reconstruction from monocular RGB videos, exploiting geometric and visual consistencies on contact maps for semi-supervised learning. To the best of our knowledge, we are the first to apply such consistencies on hand-object scenarios. We first review the literature on \emph{hand-object reconstruction}. Then, we review \emph{point cloud analysis} with the focus of graph-based methods. Finally, we provide a brief review on \emph{semi-supervised learning in 3D hand-object pose estimation}.

\subsection{Hand-object reconstruction}

Previous works mainly tackle 3D pose estimations on hands \cite{simon2017hand,zimmermann2017learning,mueller2018ganerated,spurr2018cross,yang2020seqhand,yang2020seqhand,tang2014latent} and objects \cite{labbe2020,li2018deepim,xiang2017posecnn,chen2020g2l,chen2021fs} separately. Joint reconstruction of hands and objects has been receiving increasing attention \cite{hasson2019learning,hasson2020leveraging,cao2020reconstructing,hasson20_handobjectconsist}. Hasson \etal~\cite{hasson2019learning} introduces an end-to-end model to regress MANO hand parameters jointly with object mesh vertices deformed from a sphere and incorporates contact losses which encourages contact surfaces and penalises penetrations between hand and object. A line of works \cite{tekin2019h+,doosti2020hope,hasson2020leveraging,cao2020reconstructing,hasson20_handobjectconsist,yang2021cpf,grady2021contactopt,huang2020hot} assume known object models and regress a 6DoF object pose instead. Other works focus on grasp synthesis \cite{corona2020ganhand,karunratanakul2020grasping,taheri2020grab,jiang2021hand}. In contrast, our method is in line with recent optimisation-based approaches for modelling 3D hand-object contact. ContactOpt~\cite{grady2021contactopt} proposes a contact map estimation network and a contact model to produce realistic hand-object interaction. ContactPose dataset~\cite{Brahmbhatt_2020_ECCV} is unique in capturing ground-truth thermal contact maps. However, 3D contact labels are seldom available and limited to constrained labratory settings. In this work, we treat contact maps as our primary learning target and leverage unannotated datasets.

\subsection{Point cloud analysis} 
Since point cloud data is irregular and unordered, early works tend to project the original point clouds to intermediate voxels \cite{maturana2015voxnet} or images \cite{you2018pvnet}, \ie translating into a well-explored 2D image problem. As information loss caused by projection degrades the representational quality, PointNet \cite{qi2017pointnet} is proposed to directly process unordered point sets and PointNet++ \cite{qi2017pointnet++} is extends on local point representation in multi-scale. As PointNet++ \cite{qi2017pointnet++} can be view as the generic point cloud analysis network framework, the research focus has been shifted to generating better regional points representation. Methods can be divided into convolution \cite{wu2019pointconv,xu2021paconv}, graph \cite{wang2019dynamic,li2019deepgcns,lin2020convolution} and attention \cite{guo2021pct,zhao2021point} -based. 
\noindent \textbf{Graph-based methods.}
GCNs have been gaining much attention in the last few years. This is due to two reasons: 1) the rapid increase of non-Euclidean data in real-world applications and 2) the limited performance of convolutional neural networks when dealing with such data. As the unstructured nature of point clouds poses a representational challenge in the community, graph-based methods treat points as nodes of a graph and formulate edges according to their spatial/feature relationships. MoNet \cite{monti2017geometric} defines the convolution as Gaussian mixture models in a local pseudo-coordinate system. 3D-GCN \cite{lin2020convolution} proposes a deformable kernels which has shift and scale-invariant properties for point cloud processing. DGCNN \cite{wang2019dynamic} proposes to gather nearest neighbouring points in feature space and follow by the EdgeConv operators for feature extraction. The EdgeConv operator dynamically computes node adjacency at each graph layer using the distance between point features. In this paper, we propose a computationally efficient network for contact map estimation which requires less than half the parameters of PointNet \cite{qi2017pointnet} and GPU memory of DGCNN \cite{wang2019dynamic}.

\subsection{Semi-supervised learning in 3D hand-object pose estimation}
Learning from both labelled and unlabelled data simultaneously has recently attracted growing interest
in 3D hand pose estimation~\cite{yang2021semihand,spurr2021adversarial,kaviani2021semi,chen2019so,tang2013real}.
They typically focus on training models with a small amount of labelled data as well as a relatively larger amount of unlabelled data. 
After training on human-annotated datasets, pseudo-labelling and consistency training can be used to train further and a teacher-student network with exponential moving average (EMA) strategy~\cite{wang20213dioumatch} is common to accelerate the training. 
For instance, So-HandNet~\cite{chen2019so} leverages the consistency between the recovered hand point cloud and the original hand point cloud for semi-supervised training.
SemiHand~\cite{yang2021semihand} is the first to combine pseudo-labelling and consistency learning for hand pose estimation. 
Liu \etal~\cite{liu2021semi} is the only prior work on 3D hand-object pose estimation with semi-supervised learning.
They proposed spatial and temporal constraints for selecting the pseudo-labels from videos.
However, they are limited to pseudo hand labels and did not account for physical contact with manipulated objects.
In contrast, our work is the first to explore pseudo-labelling for 3D hand-object contact map with geometric and visual consistency constraints.

\section{Methodology}
Given a noisy estimate of hand and object meshes from an image-based algorithm, we seek to learn a hand-object contact region estimator by exploiting real-world hand and object video datasets without contact ground-truths. Figure \ref{fig:highlevel_framework} shows an overview of our approach. 
In the following section, we describe our learned contact map estimation network (GCN-Contact) in Section \ref{sec:ContactNet} and our newly proposing semi-supervised training pipeline (S$^2$Contact) in Section \ref{sec:SemiContact} that utilise a teacher-student mutual learning framework.

\begin{figure}[t]
\centering
\includegraphics[width=1.0\linewidth]{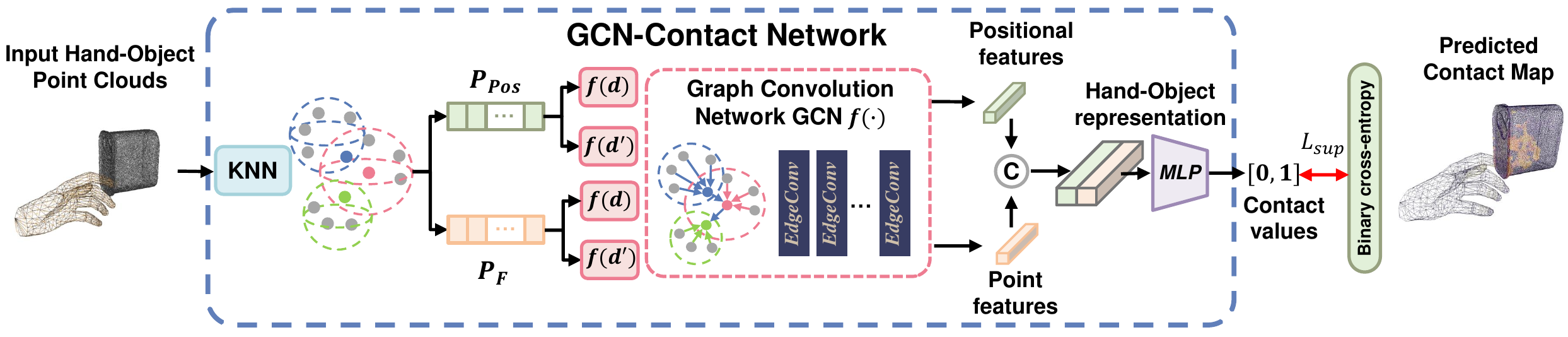}
\caption{Framework of GCN-Contact. The network takes hand-object point clouds $\mathbf{P}=(\mathbf{P}_{hand},\mathbf{P}_{obj})$ as input and perform $K$-NN search separately on 3D position $\mathbf{P}_{pos}$ and point features $\mathbf{P}_F$, \ie $\mathbf{P} = \{\mathbf{P}_{pos},\mathbf{P}_F\}$. Different dilation factors $d,d'$ are used to enlarge the receptive field for graph convolution $f(\cdot)$. Finally, features are concatenated and pass to MLP to predict contact map.
}
\vspace{-0.4cm}
\label{fig:ContactNet}
\end{figure}

\subsection{GCN-Contact: 3D hand-object contact estimation} \label{sec:ContactNet}
As pose estimates from an image-based algorithm can be potentially inaccurate, GCN-Contact learns to infer contact maps $\mathbf{C}=(\mathbf{C}_{hand},\mathbf{C}_{obj})$ from hand and object point clouds $\mathbf{P}=(\mathbf{P}_{hand},\mathbf{P}_{obj})$. We adopted the differential MANO \cite{romero2017embodied} model from \cite{hasson2019learning}. It maps pose ($\boldsymbol{\theta}\in\mathbb{R}^{51}$) and shape ($\boldsymbol{\beta}\in\mathbb{R}^{10}$) parameters to a mesh with $N=778$ vertices. Pose parameters ($\boldsymbol{\theta}$) consists of $45$ DoF (\ie $3$ DoF for each of the $15$ finger joints) plus $6$ DoF for rotation and translation of the wrist joint. Shape parameters ($\boldsymbol{\beta}$) are fixed for a given person. We sample $2048$ points randomly from object model to form object point cloud. Following \cite{grady2021contactopt}, we include $F$-dimensional point features for each point: binary per-point feature indicating whether the point belongs to the hand or object, distances from hand to object and surface normal information. With network input $\mathbf{P}=(\mathbf{P}_{hand},\mathbf{P}_{obj})$ where $\mathbf{P}_{hand}\in\mathbb{R}^{778\times F}$ and $\mathbf{P}_{obj}\in\mathbb{R}^{2048\times F}$, GCN-Contact can be trained to infer discrete contact representation ($\mathbf{C}=(\mathbf{C}_{hand},\mathbf{C}_{obj}) \in [0,1]$) \cite{Brahmbhatt_2020_ECCV} using binary cross-entropy loss. Similarly to \cite{grady2021contactopt}, the contact value range $[0,1]$ is evenly split into $10$ bins and the training loss is weighted to account for class imbalance.

\paragraph{\textbf{Revisiting PointNet-based methods.}}
Recent contact map estimators are based on PointNet \cite{jiang2021hand} and PoinetNet++ \cite{grady2021contactopt}.
PointNet \cite{qi2017pointnet} directly processes unordered point sets using shared multi-layer perceptron (MLP) networks. 
PointNet++ \cite{qi2017pointnet++} learns hierarchical features by stacking multiple learning stages and recursively capturing local geometric structures. 
At each learning stage, farthest point sampling (FPS) algorithm is used to re-sample a fixed number of points and $K$ neighbours are obtained from ball query's local neighbourhood for each sampled point to capture local structures. The kernel operation of PointNet++ for point $p_i \in\mathbb{R}^{F}$ with $F$-dimensional features can be described as:
\begin{align} \label{eq:pointnet++}
    \dot{p_i} = \sigma\big(\Phi(p_{j}| j \in \mathcal{N}(i))\big),
\end{align}
where the updated point $\dot{p_i}$ is formed by max-pooling function $\sigma(\cdot)$ and PointNet as the basic building block for local feature extractor $\Phi(\cdot)$ around point neighbourhood $\mathcal{N}(i)$ of point $p_i$. The kernel of the point convolution can be implemented with MLPs.
However, MLPs are unnecessarily performed on the neighbourhood features which
causes a considerable amount of latency in PointNet++ \cite{qian2021assanet}.
This motivates us to employ advanced local feature extractors such as convolution \cite{wu2019pointconv,xu2021paconv}, graph \cite{wang2019dynamic,li2019deepgcns,lin2020convolution} or self-attention mechanisms \cite{guo2021pct,zhao2021point}.

\paragraph{\textbf{Local geometric information.}}
While contact map estimation can take advantage of detailed local geometric information, they usually suffer from two major limitations. First, the computational complexity is largely increased with delicate extractors which leads to low inference latency. 
For instance, in graph-based methods, neighbourhood information gathering modules are placed for better modelling of the locality on point clouds. This is commonly established by $K$-nearest neighbour ($K$-NN) search which increases the computational cost quadratically with the number of points and even further for dynamic feature representation \cite{wang2019dynamic}. For reference on ModelNet$40$ point cloud classification task \cite{qian2021assanet}, the inference speed of PointNet \cite{qi2017pointnet} is $41$ times faster than DGCNN \cite{wang2019dynamic}.
Second, Liu \etal's investigation on local aggregation operators reveals that advanced local feature extractors make surprisingly similar contributions to the network performance under the same network input~\cite{liu2020closer}. For these reasons, we are encouraged to develop a computationally efficient design while maintaining comparable accuracy for learning contact map estimation.

\paragraph{\textbf{Proposed method.}}
To overcome the aforementioned limitations, we present a simple yet effective graph-based network for contact map estimation. We use EdgeConv \cite{wang2019dynamic} to generate edge features that describe the relationships between a point and its neighbours:
\begin{align} \label{eq:edgeconv}
    \Phi(p_i,p_j) = \text{ReLU}\big(\text{MLP}(p_j-p_i,p_i)\big),\;\; j\in \mathcal{N}(i),
\end{align}
where neighbourhood $\mathcal{N}(i)$ is obtained by $K$-NN search around the point $p_i$. As shown in Figure \ref{fig:ContactNet}, we only compute $K$-NN search once at each network pass to improve computational complexity and reduce memory usage. 
In addition, we apply dilation on the $K$-NN results to increase the receptive field without loss of resolution. To better construct local regions when hand and object are perturbed, we propose to perform $K$-NN search on 3D position and point features separately. Note that \cite{grady2021contactopt} perform ball query on $0.1-0.2m$ 
radius and \cite{wang2019dynamic} combine both position and features. 
Finally, we take inspirations from the Inception model \cite{szegedy2015going} in which they extract
multi-scale information by using different kernel sizes in different paths of the architecture. Similarly, we process spatial information at various dilation factors and then aggregates.
The experiment demonstrates the effectiveness of our proposed method and is able to achieve constant memory access cost regardless of the size of dilation factor $d$ (see Table \ref{table:contactpose}).

\subsection{S$^2$Contact: Semi-supervised training pipeline} \label{sec:SemiContact}
\begin{figure}[t]
\centering
\includegraphics[width=1.0\linewidth]{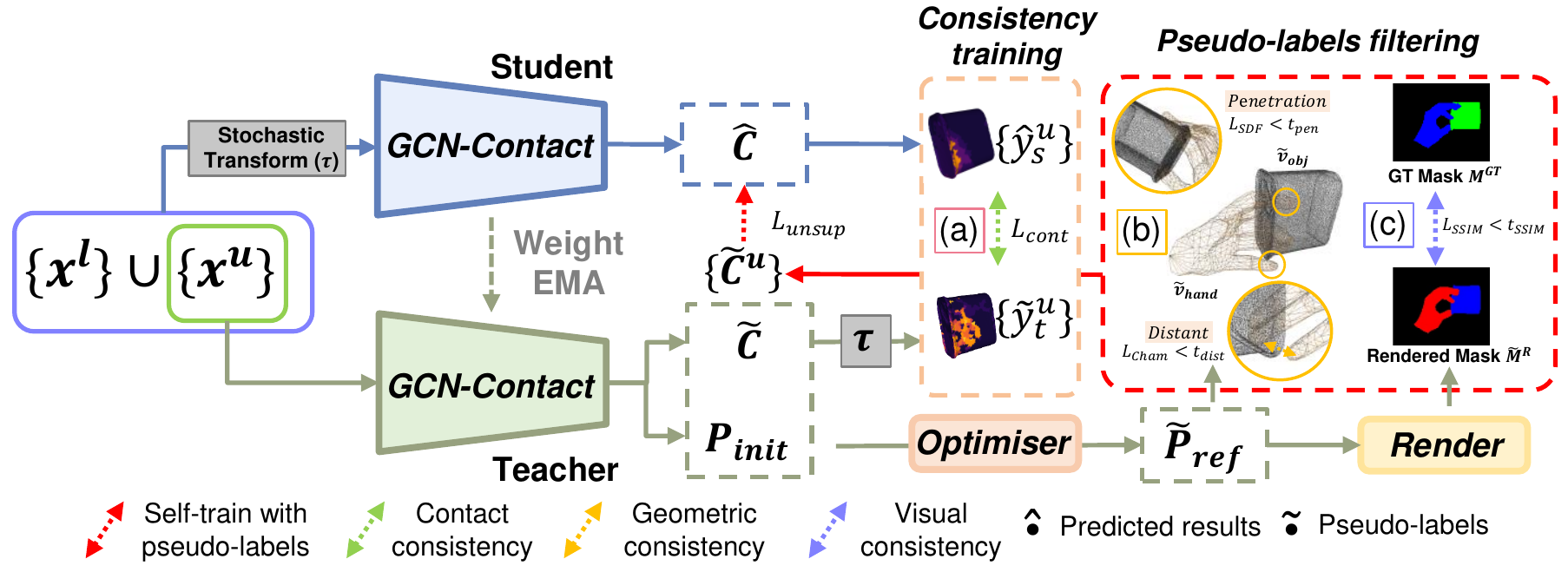}
\caption{
S$^2$Contact pipeline. We adopt our proposed graph-based network GCN-Contact as backbone. We utilise a teacher-student mutual learning framework which is composed of a learnable student and an EMA teacher. The student network is trained with labelled data $\left\{\mathbf{x}^{l},\mathbf{y}^{l}\right\}$. For unlabelled data $\mathbf{x}^{u}$, the student network takes pseudo contact labels $\widetilde{\mathbf{C}}^{u}$ from its EMA teacher and compares with its predictions $\widehat{\mathbf{C}}$. Please note that pseudo contact labels $\widetilde{\mathbf{C}}^{u}$ is a subset of $\widetilde{\mathbf{C}}$, \ie $\widetilde{\mathbf{C}}^{u} \in \widetilde{\mathbf{C}}$. 
(a) refers to contact consistency constraint for consistency training.
To improve the quality of pseudo-label, we adopt a confidence-based filtering mechanism to geometrically (b) and visually (c) filter out predictions that violate contact constraints.
}
\label{fig:SemiContact}
\end{figure}
Collecting ground-truth contact annotation for hand-object dataset can be both challenging and time-consuming.
To alleviate this, we introduce a semi-supervised learning framework to learn 3D hand-object contact estimation by leveraging large-scale unlabelled videos.
As shown in Figure~\ref{fig:highlevel_framework}, our proposed framework relies on two training stages: 1) pre-training stage where the model is pre-trained on the existing labelled data~\cite{Brahmbhatt_2020_ECCV}; 2) semi-supervised stage where the model is trained by the pseudo-labels from  unlabelled hand-object datasets~\cite{chao2021dexycb,hampali2020honnotate,FirstPersonAction_CVPR2018}. 
As pseudo-labels are often noisy, we propose confidence-based filtering with geometric and visual consistency constraints to improve the quality of pseudo-labels.

\paragraph{\textbf{Pre-training.}} 
As good initial contact estimate enables semi-supervision, we pre-train our graph-based contact estimator using a small labelled dataset $\left\{\mathbf{x}^{l},\mathbf{y}^{l}\right\}$. We followed \cite{grady2021contactopt} and optimise hand-object poses to achieve target contact. Upon convergence, we clone the network to create a pair of student-teacher networks.

\paragraph{\textbf{Pseudo-label generation.}}
To maintain a reliable performance margin over the student network throughout the training, we adopt an EMA teacher which is commonly used in semi-supervised learning. The output of the student network is the predicted contact map $\widehat{\mathbf{C}}$.
The teacher network generates pseudo-labels which includes pre-filter contact map $\widetilde{\mathbf{C}}$ and refined hand-object pose $\widetilde{P}_{ref}$. As it is crucial for the teacher network to generate high-quality pseudo-labels under a semi-supervised framework, we propose a confidence-based filtering mechanism that leverages geometric and visual consistency constraints. 

\paragraph{\textbf{Contact consistency constraint for consistency training.}}
We propose a contact consistency loss to encourage robust and stable predictions for unlabelled data $\mathbf{x}^{u}$.
As shown in Figure~\ref{fig:SemiContact} (a), we first apply stochastic transformations $\mathcal{T}$ which includes flipping, rotation and scaling on the input hand-object point clouds $\mathbf{x}^{u}$ for the student network. 
The predictions of the student network $\widehat{\mathbf{y}}^{u}_{s} \in \widehat{\mathbf{C}}$ are compared with the teacher predictions $\widetilde{\mathbf{y}}^{u}_{t} \in \widetilde{\mathbf{C}}$ processed by the same transformation $\mathcal{T}$ using contact consistency loss: 
\begin{align}
\begin{split}
\mathcal{L}_{cont}(\mathbf{x}^{u}) &=\|\Omega(\mathcal{T}(\mathbf{x}^{u}))-(\mathcal{T}(\Omega(\mathbf{x}^{u})))\|_{1} \\
&=\|\widehat{\mathbf{y}}^{u}_{s}-\widetilde{\mathbf{y}}^{u}_{t}\|_{1},
\end{split}
\end{align}
where $\Omega(\cdot)$ represents the predicted contact map.

\paragraph{\textbf{Geometric consistency constraint for pseudo-labels filtering.}}
As shown in Figure~\ref{fig:SemiContact} (b), we propose a geometric consistency constraint to the hand and object pseudo pose label $\widetilde{P}_{ref}$.
Concretely, we allow the Chamfer distance $\mathcal{L}_{Cham}$ between hand and object meshes to be less than threshold $t_{dist}$:
\begin{align}
    \mathcal{L}_{Cham}(\widetilde{\mathbf{v}}_{hand},\widetilde{\mathbf{v}}_{obj}) = &\frac{1}{\left|\widetilde{\mathbf{v}}_{obj}\right|}\sum_{x \in \widetilde{\mathbf{v}}_{obj}}d_{\widetilde{\mathbf{v}}_{hand}}(x) + \frac{1}{\left|\widetilde{\mathbf{v}}_{hand}\right|}\sum_{y \in \widetilde{\mathbf{v}}_{hand}}d_{\widetilde{\mathbf{v}}_{obj}}(y), 
\end{align}
where $\widetilde{\mathbf{v}}_{hand}$ and $\widetilde{\mathbf{v}}_{obj}$ refers to hand and object point sets, $d_{\widetilde{\mathbf{v}}_{hand}}(x)=\min_{y \in \widetilde{\mathbf{v}}_{hand}}\allowbreak\left\|x-y\right\|_{2}^{2}$, and $d_{\widetilde{\mathbf{v}}_{obj}}(y)=\min_{x \in \widetilde{\mathbf{v}}_{obj}}\left\|x-y\right\|_{2}^{2}$.
Similarly for interpenetration, we use $\mathcal{L}_{SDF}(\widetilde{\mathbf{v}}_{obj})=\sum_{hand,obj} \sum_{i} \Psi_{h}\left(\widetilde{\mathbf{v}}_{obj}^{i}\right) \leq t_{pen}$ to ensure object is being manipulated by hand.
$\Psi_{h}$ is the Signed Distance Field (SDF) from the
hand mesh (\ie, $\Psi_{h}(\widetilde{\mathbf{v}}_{obj})=-\min \left(\operatorname{SDF}\left(\widetilde{\mathbf{v}}_{obj}\right), 0\right)$) to detect object penetrations.

\paragraph{\textbf{Visual consistency constraint for pseudo-labels filtering.}} 
We observed that geometric consistency is insufficient to correct hand grasp (see Table \ref{table:ab_ssl}). 
To address this, we propose a visual consistency constraint to filter out the pseudo-labels whose rendered hand-object image $\widetilde{I}_{ho}$ does not match the input image. 
We first use a renderer~\cite{kato2018neural} to render the hand-object image from the refined pose $\widetilde{P}_{ref}$ and obtain the hand-object segment of the input image $I$ by applying the segmentation mask $M_{gt}$. 
Then, the structural similarity (SSIM)~\cite{wang2004image} between two images can be computed. We keep pseudo-labels when $\mathcal{L}_{SSIM} \leq t_{SSIM}$:
\begin{equation}
\mathcal{L}_{SSIM}(I,M_{gt},\widetilde{I}_{ho})=1-SSIM\left(I \odot M_{gt}, \widetilde{I}_{ho}\right),
\end{equation}
where $\odot$ denotes element-wise multiplication. 

\paragraph{\textbf{Self-training with pseudo-labels.}} 
After filtering pseudo-labels, our model is trained with the union set of the human-annotated dataset and the remaining pseudo-labels.
The total loss $\mathcal{L}_{semi}$ can be described as:
\begin{equation}
\mathcal{L}_{semi}(\widehat{\mathbf{C}},\mathbf{y}^{l},\widetilde{\mathbf{C}}^{u},\mathbf{x}^{u})=\mathcal{L}_{sup}\left(\widehat{\mathbf{C}},\mathbf{y}^{l}\right)+\mathcal{L}_{unsup}\left(\widehat{\mathbf{C}},\widetilde{\mathbf{C}}^{u}\right)+\lambda_{c} \mathcal{L}_{cont}\left(\mathbf{x}^{u}\right),
\end{equation}
where $\mathcal{L}_{sup}$ is a supervised contact loss, $\mathcal{L}_{unsup}$ is a unsupervised contact loss with pseudo-labels and $\lambda_{c}$ is a hyperparameter. Note that $\mathcal{L}_{sup}$ (see Figure \ref{fig:ContactNet}) and $\mathcal{L}_{unsup}$ (see Figure \ref{fig:SemiContact}) are both binary cross-entropy loss.

\section{Experiments}

\paragraph{\textbf{Implementation details.}}
We implement our method in PyTorch~\cite{PyTorch}. All experiments are run on an Intel i9-CPU @ 3.50GHZ, 16 GB RAM, and one NVIDIA RTX 3090 GPU. 
For pseudo-labels filtering, $t_{dist}=0.7$, $t_{pen}=6$ and $t_{SSIM}=0.25$ are the constant thresholds and stochastic transformations includes flipping ($\pm20\%$), rotation ($\pm180^{\circ}$) and scaling ($\pm20\%$). We train all parts of the network simultaneously with Adam optimiser \cite{kingma2014adam} at a learning rate $10^{-3}$ for 100 epochs. We empirically fixed $K=10,d=4$ to produce the best results.

\paragraph{\textbf{Datasets and evaluation metrics.}}
\emph{ContactPose} is the first dataset \cite{Brahmbhatt_2020_ECCV} of hand-object contact paired with hand pose, object pose and RGB-D images. It contains 2,306 unique grasps of $25$ household objects grasped with $2$ functional intents by $50$ participants, and more than $2.9$M RGB-D grasp images. For fair comparisons with ContactOpt \cite{grady2021contactopt}, we follow their \emph{Perturbed ContactPose} dataset where hand meshes are modified by additional noise to MANO parameters. This results in 22,624 training and 1,416 testing grasps. \emph{DexYCB} is a recent real dataset for capturing hand grasping of objects \cite{chao2021dexycb}. It consists of a total of 582,000 image frames on 20 objects from the YCB-Video dataset \cite{xiang2017posecnn}. We present results on their default official dataset split settings. \emph{HO-3D}~\cite{hampali2020honnotate} is similar to \emph{DexYCB} where it consists of 78,000 images frames on 10 objects. We present results on the official dataset split (version 2). The hand mesh error is reported after procrustes alignment and in $mm$.

\begin{itemize}
\item \emph{Hand error:} We report the mean end-point error ($mm$) over 21 joints and mesh error in $mm$.
\item \emph{Object error:} We report the percentage of average object 3D vertices error within $10\%$ of object diameter (ADD-$0.1$D).
\item \emph{Hand-object interaction:} 
We report the intersection volume ($cm^{3}$) and contact coverage ($\%$). Intersection volume is obtained by voxelising the hand and object using a voxel size of $0.5cm$. Contact coverage refers to the percentage of hand points between $\pm2mm\%$ of the object surface~\cite{grady2021contactopt}.
\end{itemize}

\paragraph{\textbf{Baseline.}}
For refining image-based pose estimates, we use the baseline pose estimation network from Hasson \etal~\cite{hasson2020leveraging} and retrain it on the training split of the respecting dataset. We filter out frames where the minimum distance between the ground truth hand and object surfaces is greater than 2 $mm$. We also use the contact estimation network DeepContact from Grady \etal~\cite{grady2021contactopt} which takes ground-truth object class and pose. For semi-supervised learning, we use the baseline method from Liu \etal~\cite{liu2021semi}, a semi-supervised learning pipeline for 3D hand-object pose estimation from large-scale hand-object interaction videos.

\subsection{Comparative results}
\begin{figure}[t]
\centering
\includegraphics[width=0.8\linewidth]{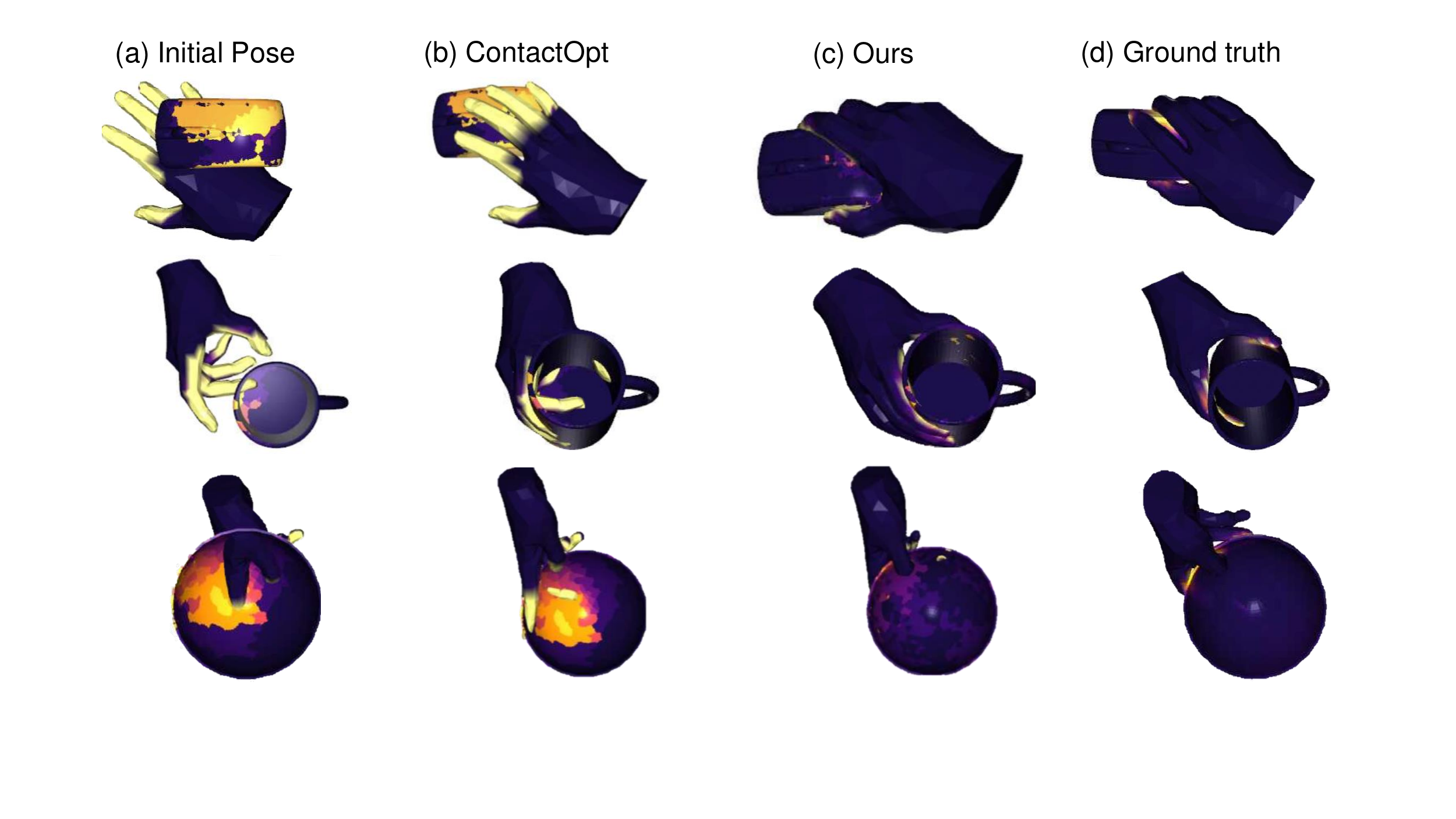}
\caption{Qualitative comparison with ContactOpt~\cite{grady2021contactopt} on \emph{ContactPose}. We observed that penetrations across hand and object (can be seen in (b)) is likely to be caused by contact predictions appearing on both object surfaces. Our model, trained only on \emph{ContactPose}, shows robustness to various hand poses and objects.
}
\label{fig:vis_contactpose}
\end{figure}

\paragraph{\textbf{Refining small and large inaccuracies.}}
We use \emph{ContactPose} to evaluate GCN-Contact for refining poses with small (\emph{ContactPose}) and large (\emph{Perturbed ContactPose}) inaccuracies. Table \ref{table:contactpose} shows the results for both cases. For \emph{Perturbed ContactPose}, the mean end-point error over $21$ joints is $82.947mm$ before refinement. This is aimed at testing the ability to improve hand poses with large errors. In contrast, \emph{ContactPose} is used to evaluate $mm$-scale refinement. As shown, our method consistently outperforms baseline and DGCNN \cite{wang2019dynamic}. We attribute the performance gain to multi-scale feature aggregation with dilation. Qualitative comparison with ContactOpt~\cite{grady2021contactopt} is provided in Figure \ref{fig:vis_contactpose}.

\newcolumntype{C}{>{\centering\arraybackslash}X}
\begin{table}[t]
\begin{center}
\caption{Hand error rates ($mm$) on \emph{Perturbed ContactPose} and \emph{ContactPose} datasets.}
\label{table:contactpose}
\resizebox{0.85\linewidth}{!}{%
\begin{tabularx}{\linewidth}{l | C  C | C  C | C  C }
\toprule
 & \multicolumn{2}{c|}{Baseline} & \multicolumn{2}{c|}{DGCNN \cite{wang2019dynamic}} & \multicolumn{2}{c}{Ours} \\ 
 & joint $\downarrow$ & mesh $\downarrow$ & joint $\downarrow$ & mesh $\downarrow$ & joint $\downarrow$ & mesh $\downarrow$\\
\midrule 
\emph{Perturbed ContactPose} & {\footnotesize32.988} &{\footnotesize33.147} & {\footnotesize32.592} & {\footnotesize32.762} & \textbf{{\footnotesize29.442}} & \textbf{{\footnotesize30.635}}\\
\emph{ContactPose}  & {\footnotesize8.880} & {\footnotesize8.769} & {\footnotesize8.767} & {\footnotesize32.988} & \textbf{{\footnotesize5.878}} & \textbf{{\footnotesize5.765}}\\
\bottomrule
\end{tabularx}
}
\end{center}
\end{table}

\begin{figure}[t]
\centering
{\includegraphics[width=0.85\linewidth]{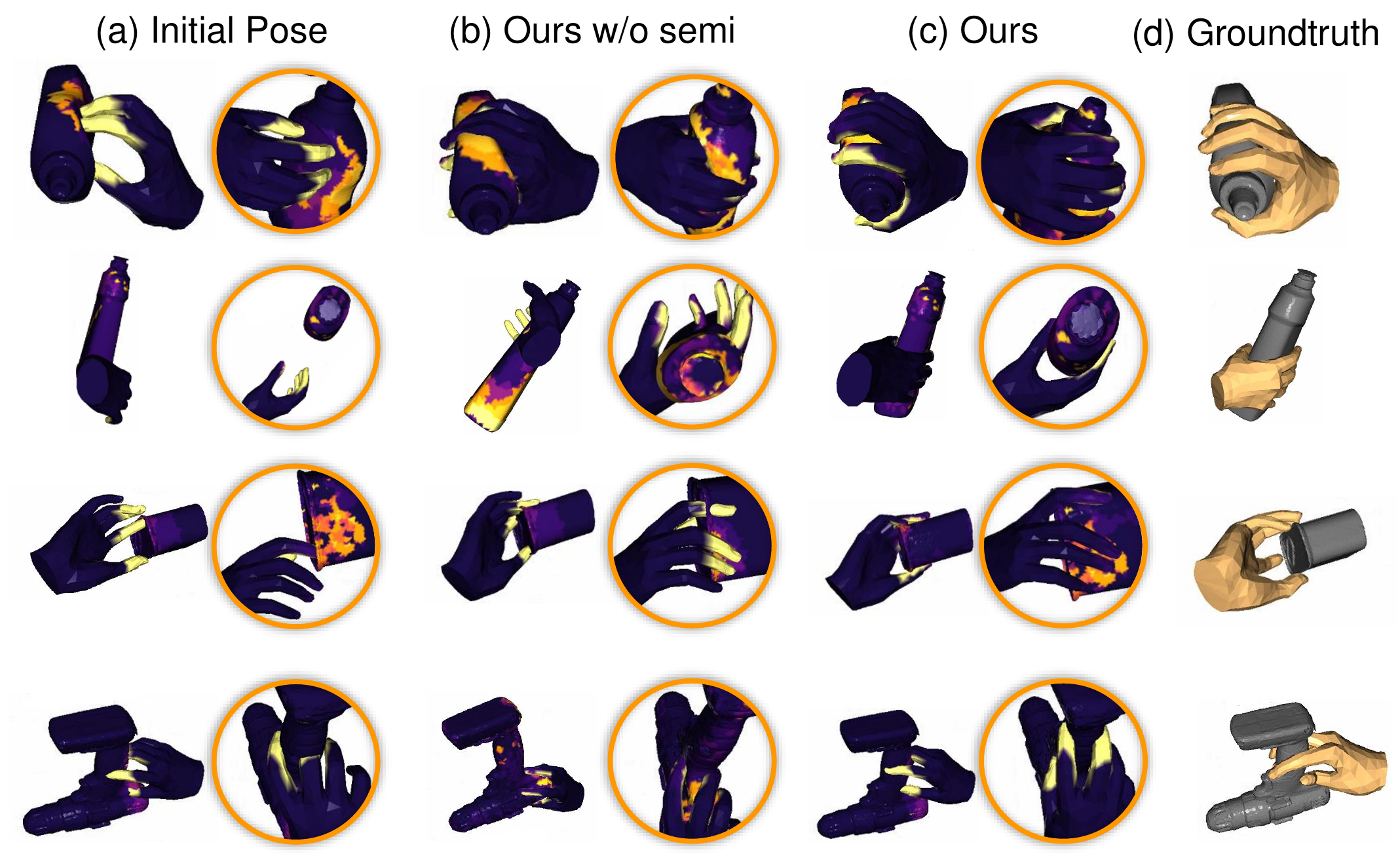}}
\caption{Qualitative comparison with Initial Pose~\cite{liu2021semi} on \emph{HO-3D}. (a) We observed that the reconstructions of hand and object are more likely to be distant in 3D space due to depth and scale ambiguity with only RGB image as an input. (b) In contrast, when without semi-supervised learning, there exists more interpenetration or incorrect grasps between hand and object. (c) As shown, together with visual cues and contact map estimations, our method is able correct the above failure cases and generate more physically-plausible reconstructions.
}
\label{fig:vis_ho3d}
\end{figure}

\paragraph{\textbf{Refining Image-based pose estimates.}}
We evaluate S$^2$Contact in refining poses from an image-based pose estimator. We use the baseline image-based pose estimation network from Hasson \etal~\cite{hasson2020leveraging} and retrain it on the training split of the respecting dataset. Unlike \cite{grady2021contactopt}, we do not rely on ground-truth object class and pose. 
In particular, we compare with the current state-of-the-art~\cite{liu2021semi} which is also a semi-supervised framework for 3D hand-object pose estimation. 
Liu \etal~\cite{liu2021semi} proposes spatial-temporal consistency in large-scale hand-object videos to generate pseudo-labels for hand. In contrast, we leverage physical contact and visual consistency constraints to generate pseudo contact labels which can be optimised jointly with hand and object poses. 
As shown in Table~\ref{table:semisupervised}, our method outperforms \cite{liu2021semi} by $11.5\%$ in average object ADD-$0.1$D score.
Besides, we also compare with our baseline contact model ContactOpt~\cite{grady2021contactopt}.
As shown in  Table~\ref{table:semisupervised}, we are able to further improve hand error by $1mm$ and $0.8mm$ over joints and mesh. 
In addition to hand-object pose performance, our method is able to better reconstruct hand and object with less intersection volume and higher contact coverage.
The above demonstrates that our method provides a more practical alternative to alleviate the reliance on heavy dataset annotation in hand-object. In addition, we provide qualitative comparison on \emph{HO-3D} in Figure \ref{fig:vis_ho3d}. We also report the cross-dataset generalisation performance of our model
on \emph{DexYCB} in Table~\ref{table:crossdataset}. We select three objects (\ie, mustard bottle, potted meat can and bleach cleanser), to be consistent with \emph{HO-3D}. 
As shown, our method consistently shown improvements across all metrics.

\newcolumntype{C}{>{\centering\arraybackslash}X}
\begin{table}[t]
\begin{center}
\caption{
Error rates on \emph{HO-3D}. Note that Liu \etal~\cite{liu2021semi} is the current state-of-the-art semi-supervised method. \textbf{ave}, \textbf{inter} and \textbf{cover} refers to average, intersection volume and contact coverage, respectively.
}
\label{table:semisupervised}
\resizebox{0.9\linewidth}{!}{%
\begin{tabularx}{\linewidth}{l | C  C | C  C C C | C  c }
\toprule
 & \multicolumn{2}{c|}{Hand error} & \multicolumn{4}{c|}{Object ADD-0.1D($\uparrow$)} & \multicolumn{2}{c}{Contact} \\ 
Methods & joint $\downarrow$ & mesh $\downarrow$ & bottle & can & bleach & avg & cover $\uparrow$ & inter $\downarrow$\\
\midrule 
Initial Pose~\cite{hasson2020leveraging} & {\footnotesize11.1} &{\footnotesize11.0} & {\footnotesize-} & {\footnotesize-} & {\footnotesize-} & {\footnotesize74.5} & {\footnotesize4.4} & {\footnotesize15.3$\pm$21.1}\\
ContactOpt~\cite{grady2021contactopt}  & {\footnotesize9.7} & {\footnotesize9.7} & {\footnotesize-} & {\footnotesize-} & {\footnotesize-} & {\footnotesize75.5} & {\footnotesize14.7} & {\footnotesize6.0$\pm$6.7}\\
\midrule 
Liu \etal~\cite{liu2021semi} & {\footnotesize9.9} &{\footnotesize9.5} & {\footnotesize69.6} & {\footnotesize53.2} & {\footnotesize86.9} & {\footnotesize69.9} & {\footnotesize-} & {\footnotesize-}\\
Ours & \textbf{{\footnotesize8.7}} & \textbf{{\footnotesize8.9}} & \textbf{{\footnotesize79.1}} & \textbf{{\footnotesize71.8}} & \textbf{{\footnotesize93.3}} & \textbf{{\footnotesize81.4}} & \textbf{{\footnotesize19.2}} &\textbf{{\footnotesize 3.5$\pm$1.8}}\\
\bottomrule
\end{tabularx}
}
\end{center}
\end{table}
\newcolumntype{C}{>{\centering\arraybackslash}X}
\begin{table}[t]
\begin{center}
\caption{
Error rates of the cross-dataset generalisation performance on \emph{DexYCB}.
}
\label{table:crossdataset}
\resizebox{0.85\linewidth}{!}{%
\begin{tabularx}{\linewidth}{l | C  C | C  C C C | C  c }
\toprule
 & \multicolumn{2}{c|}{Hand error} & \multicolumn{4}{c|}{Object ADD-0.1D($\uparrow$)} & \multicolumn{2}{c}{Contact} \\ 
models & joint $\downarrow$ & mesh $\downarrow$ & bottle & can & bleach & avg & cover $\uparrow$ & inter $\downarrow$\\
\midrule 
w/o semi-supervised & {\footnotesize13.0} &{\footnotesize13.7} & {\footnotesize76.9} & {\footnotesize46.2} & {\footnotesize68.4} & {\footnotesize63.8} & {\footnotesize4.1} & {\footnotesize16.0$\pm$12.3}\\
semi-supervised  & \textbf{{\footnotesize11.8}} & \textbf{{\footnotesize12.1}} & \textbf{{\footnotesize83.6}} & \textbf{{\footnotesize53.7}} &\textbf{{\footnotesize74.2}} & \textbf{{\footnotesize70.5}} & \textbf{{\footnotesize9.3}} & \textbf{{\footnotesize10.5$\pm$7.9}}\\
\bottomrule
\end{tabularx}
}
\end{center}
\end{table}

\subsection{Ablation study}
\paragraph{\textbf{Number of $K$ neighbours.}}
Table \ref{table:ab_graph}(a) shows the results of varying number of $K$ neighbours without dilation. As shown, increasing $K$ improves immediately at $K=10$ but does not gain performance further. 
\vspace{-0.2cm}
\paragraph{\textbf{Size of dilation factor $d$.}}
As shown above that performance saturates at $K=10$, we now fix $K$ and vary the size of dilation factor $d$ in Table \ref{table:ab_graph}(b). We find that the combination $K=10, d=4$ produce the best performance and do not improve by further increasing $d$. This demonstrates the effectiveness of increasing the receptive field for contact map estimation.
\vspace{-0.2cm}
\paragraph{\textbf{Combining $K$-NN computation.}}
To further study the effect of separately computing $K$-NN, we experiment with combined $K$-NN computation with $d=4$ in Table \ref{table:ab_graph}(c). It can be seen that the performance exceed the lower bound of (a) and similar to DGCNN's performance in Table \ref{table:contactpose}. This is expected as this is similar to static EdgeConv \cite{wang2019dynamic} with dilation. This shows that separate $K$-NN is crucial for this framework.
\newcolumntype{C}{>{\centering\arraybackslash}X}
\begin{table}[t]
\begin{center}
\caption{
Performances of different GCN-Contact design choices on \emph{ContactPose} and \emph{HO-3D}. \textbf{semi} refers to semi-supervised learning. We experiment on (a) number of $K$ neighbours without dilation, (b) size of dilation factor $d$ with $K=10$ and (c) combining $K$-NN computation (denoted with *) with $d=4$. Full results are in supplementary.
}
\label{table:ab_graph}
\resizebox{0.95\linewidth}{!}{%
\begin{tabularx}{\linewidth}{l l | C  C | C  C  C }
\toprule
& & \multicolumn{2}{c|}{\emph{ContactPose}} & \multicolumn{3}{c}{\emph{HO-3D} w/o semi} \\
& & \multicolumn{2}{c|}{Hand error} & \multicolumn{2}{c}{Hand error} & {Object error}  \\ 
& models & joint $\downarrow$ & mesh $\downarrow$ & joint $\downarrow$ & mesh $\downarrow$ & add-$0.1$d ($\uparrow$) \\
\midrule 
& $K=5$ & {\footnotesize8.252} &{\footnotesize8.134}  & {\footnotesize12.69} & {\footnotesize12.86} & {\footnotesize66.33}\\
(a) & $K=10$ & \textbf{{\footnotesize6.691}} & \textbf{{\footnotesize6.562}}  & \textbf{{\footnotesize11.81}} & \textbf{{\footnotesize11.91}} & \textbf{{\footnotesize68.71}}\\
 & $K=15$ & {\footnotesize6.715} &{\footnotesize6.617}  & {\footnotesize11.85} & {\footnotesize11.92} & {\footnotesize68.70}\\
\midrule 
& $d=2$ & {\footnotesize5.959} &{\footnotesize5.865}  & {\footnotesize10.91} & {\footnotesize10.86} & {\footnotesize70.25}\\
(b) & $d=4$ & \textbf{{\footnotesize5.878}} & \textbf{{\footnotesize5.765}}  & \textbf{{\footnotesize9.92}} & \textbf{{\footnotesize9.79}} & \textbf{{\footnotesize72.81}}\\
& $d=6$ & {\footnotesize5.911} &{\footnotesize5.805}  & {\footnotesize9.95} & {\footnotesize9.79} & {\footnotesize72.81}\\
\midrule 
& $K^*=5$ & {\footnotesize8.451} &{\footnotesize8.369}  & {\footnotesize12.91} & {\footnotesize12.86} & {\footnotesize68.86}\\
(c) & $K^*=10$ & \textbf{{\footnotesize8.359}} & \textbf{{\footnotesize8.251}}  & \textbf{{\footnotesize11.55}} & \textbf{{\footnotesize11.97}} & \textbf{{\footnotesize69.10}}\\
& $K^*=25$ & {\footnotesize8.369} &{\footnotesize8.286}  & {\footnotesize11.57} & {\footnotesize11.97} & {\footnotesize69.06}\\
\bottomrule
\end{tabularx}
}
\end{center}
\end{table}
\paragraph{\textbf{Impact of our components.}}
We study the impact of semi-supervised learning on \emph{HO-3D}. Since the hand model (MANO) is consistent across datasets, the contact estimator can easily transfer hand contact to new datasets without re-training. 
However, it is insufficient to adapt to unlabelled dataset due to diverse object geometries.
Therefore, we propose a semi-supervised learning method to generate high-quality pseudo-labels. 
As shown in Table~\ref{table:ab_ssl}, our method enables performance boost on both hand and object. 
The hand joint error is $8.74mm$ while it is $9.92mm$ without semi-supervised training. 
Also, the average object ADD-$0.1$D has a significant $8.56\%$ improvement under S$^2$Contact.

Table~\ref{table:ab_ssl} shows a quantitative comparison of S$^2$Contact with various filtering constraints disabled demonstrating that constraints from both visual and geometry domains are essential for faithful training. 
We also observed that disabling $\mathcal{L}_{cont}$ can easily lead to unstable training and $5.45\%$ performance degradation in object error. 
In contrast, geometric consistencies ($\mathcal{L}_{Cham}$ and $\mathcal{L}_{SDF}$) have a comparably smaller impact on hand and object pose.
Despite that they account for less than $2\%$ performance drop to object error, geometric consistencies are important for contact (\ie, more than $5\%$ for contact coverage).
The remaining factor, measuring visual similarity, has a more significant impact. 
Disabling visual consistency constraint $\mathcal{L}_{SSIM}$ results in hand joint error and object error increase by $1mm$ and $8.04\%$, respectively.
We validate that the combination of our pseudo-label filtering constraints are critical for generating high-quality pseudo-labels and improving hand-object pose estimation performance. Finally, we provide qualitative examples on out-of-domain objects in supplementary.

\newcolumntype{C}{>{\centering\arraybackslash}X}
\begin{table}[t]
\begin{center}
\caption{
Performances of different filtering constraints under semi-supervised learning on \emph{HO-3D}. \textbf{semi} refers to semi-supervised learning.
}
\label{table:ab_ssl}
\resizebox{0.95\linewidth}{!}{%
\begin{tabularx}{\linewidth}{l | C  C | C | C  C }
\toprule
 & \multicolumn{2}{c|}{Hand error} & {Object error} & \multicolumn{2}{c}{Contact} \\ 
models & joint $\downarrow$ & mesh $\downarrow$ & add-0.1d ($\uparrow$) & cover $\uparrow$ & inter $\downarrow$\\
\midrule 
w/o semi & {\footnotesize9.92} &{\footnotesize9.79}  & {\footnotesize72.81} & {\footnotesize12.1} & {\footnotesize8.3$\pm$10.5}\\
w/ semi  & \textbf{{\footnotesize8.74}} & \textbf{{\footnotesize8.86}} & \textbf{{\footnotesize81.37}} & \textbf{{\footnotesize19.2}} &\textbf{{\footnotesize 3.5$\pm$1.8}}\\
\midrule 
w/o $\mathcal{L}_{Cham}$ & {\footnotesize8.53} &{\footnotesize8.48} & {\footnotesize80.11} & {\footnotesize13.1} & {\footnotesize6.9$\pm$6.2}\\
w/o $\mathcal{L}_{SDF}$ & {\footnotesize8.61} &{\footnotesize8.59} & {\footnotesize80.90} & {\footnotesize14.7} & {\footnotesize5.5$\pm$6.0}\\
w/o $\mathcal{L}_{cont}$ & {\footnotesize9.28} &{\footnotesize9.19} & {\footnotesize75.92} & {\footnotesize13.9} & {\footnotesize4.8$\pm$3.1}\\
w/o $\mathcal{L}_{SSIM}$ & {\footnotesize9.71} & {\footnotesize9.57} & {\footnotesize73.33} & {\footnotesize16.3} &{\footnotesize 3.7$\pm$2.6}\\ 
\bottomrule
\end{tabularx}
}
\end{center}
\end{table}

\paragraph{\textbf{Computational analysis.}}
We report the model parameters and GPU memory cost in Table 4 of supplementary material. For fair comparisons, all models are tested using a batch size of $64$. As shown, our model has $2.4$X less the number of learnable model parameters and $2$X less the GPU memory cost when compared to baseline and DGCNN \cite{wang2019dynamic}, respectively. We alleviate the need to keep a high density of points across the network (DGCNN) while gaining performance.
\section{Conclusion}
In this paper, we have proposed a novel semi-supervised learning framework which enables learning contact with monocular videos.
The main idea behind this study was to demonstrate that this can successfully be achieved with visual and geometric consistency constraints for pseudo-label generation. 
We designed an efficient graph-based network for inferring contact maps and shown benefits of combining visual cues and contact consistency constraints to produce more physically-plausible reconstructions.
In the future, we would like to explore more consistencies over time and or multiple views to further improve the accuracy.\\

\footnotesize\noindent \textbf{Acknowledgements.} This research was supported by the MSIT (Ministry of Science and ICT), Korea, under the ITRC (Information Technology Research Center) support program (IITP--2022--2020--0--01789) supervised by the IITP (Institute of Information \& Communications Technology Planning \& Evaluation) and the Baskerville Tier 2 HPC service (https://www.baskerville.ac.uk/) funded by the Engineering and Physical Sciences
Research Council (EPSRC) and UKRI through the World Class Labs scheme (EP/T022221/1) and the Digital Research Infrastructure programme (EP/W032244/1) operated by Advanced Research Computing at the University of Birmingham. KIK was supported by the National Research Foundation of Korea (NRF) grant (No. 2021R1A2C2012195) and IITP grants (IITP--2021--0--02068 and IITP--2020--0--01336). ZQZ was supported by China Scholarship Council (CSC) Grant No. 202208060266. AL was supported in part by the EPSRC (grant number EP/S032487/1). FZ was supported by the National Natural Science Foundation of China under Grant No. 61972188 and 62122035.



\clearpage
%
%

\bibliographystyle{splncs04}
\bibliography{short,cvpr2019,cvpr2020}

\end{document}